\documentclass[sigconf]{acmart}

\copyrightyear{2024}
\acmYear{2024}
\setcopyright{rightsretained}
\acmConference[MM '24]{Proceedings of the 32nd ACM International Conference on Multimedia}{October 28-November 1, 2024}{Melbourne, VIC, Australia}
\acmBooktitle{Proceedings of the 32nd ACM International Conference on Multimedia (MM '24), October 28-November 1, 2024, Melbourne, VIC, Australia}
\acmDOI{10.1145/3664647.3681618}
\acmISBN{979-8-4007-0686-8/24/10}

\makeatletter
\usepackage{amsfonts}
\usepackage{multirow}
\usepackage{rotating}

\usepackage{framed}
\newcommand{\VarSty}[1]{\textnormal{\ttfamily\color{blue!90!black}#1}\unskip}
\usepackage{colortbl} 
\usepackage{natbib}
\usepackage{enumitem} 
\usepackage{subcaption}
\usepackage{hyperref}
\usepackage{cleveref} 
\usepackage{xspace}
\usepackage{soul}
\makeatletter
\newlength\@framed@hsep
\newsavebox\@framed@box
\newenvironment{customframed}[1][]
  {\par\noindent
   \begin{lrbox}{\@framed@box}
   \begin{minipage}{\linewidth}
   \setlength{\fboxsep}{\@framed@hsep}
   \begin{framed}#1}
  {\end{framed}
   \end{minipage}
   \end{lrbox}%
   \colorbox[gray]{0.975}{\usebox{\@framed@box}}}
\makeatother

\newcommand{\etc}{\emph{etc}.\xspace}

\makeatletter
\def\hlinew#1{%
  \noalign{\ifnum0=`}\fi\hrule \@height #1 \futurelet
   \reserved@a\@xhline}
\makeatother

\begin{document}

\title{
    \raisebox{-0.35ex}{\includegraphics[height=1.5em]{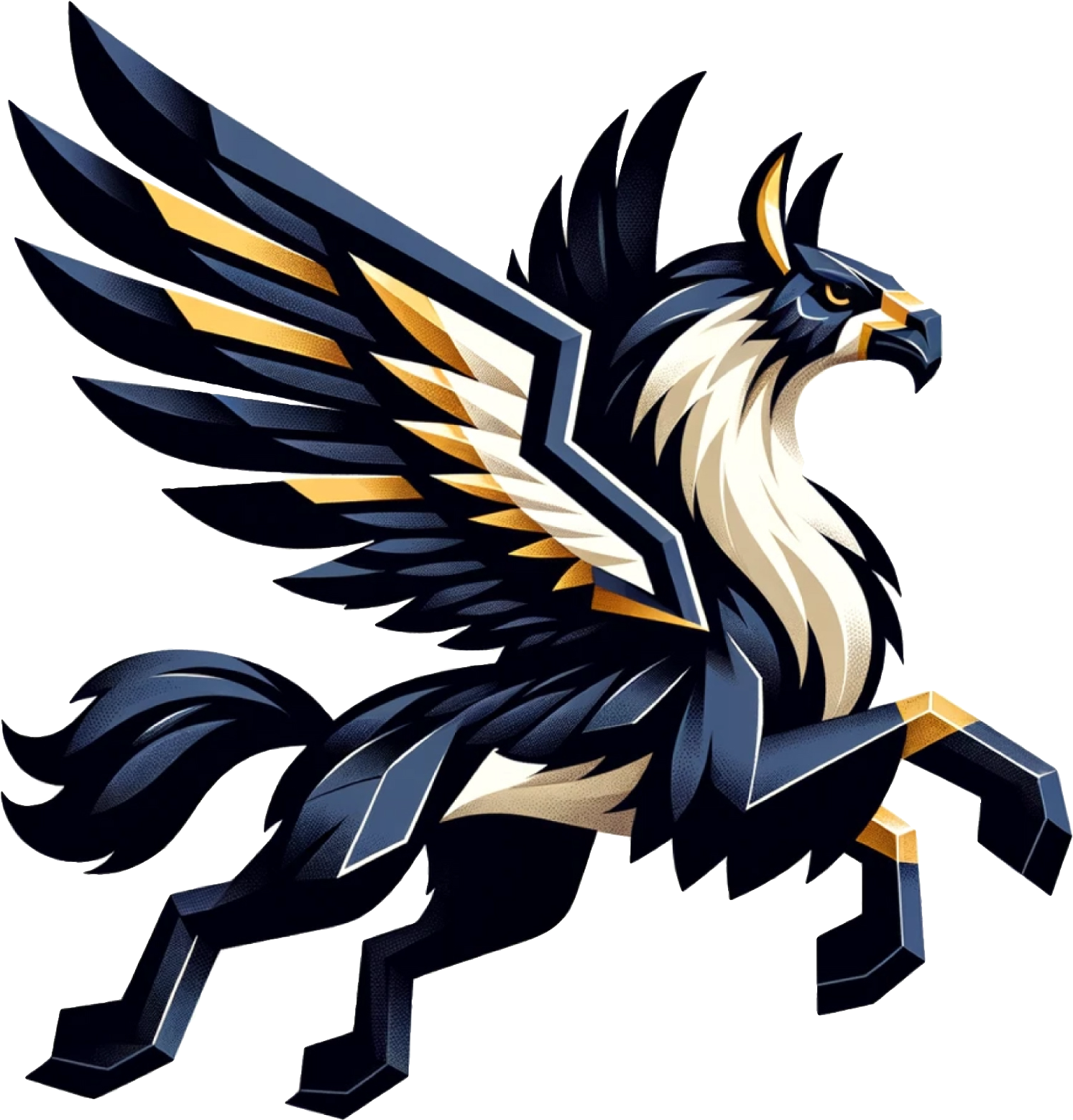}}\hspace{0.3em}EAGLE: Egocentric AGgregated Language-video Engine
}





\author{Jing Bi}
\email{jing.bi@rochester}
\orcid{0009-0006-8235-2158}
\affiliation{%
  \institution{University of Rochester}
  \city{}
  \state{}
  \country{}
}

\author{Yunlong Tang}
\email{yunlong.tang@rochester.edu}
\affiliation{\institution{University of Rochester}
  \city{}
  \state{}
  \country{}
}

\author{Luchuan Song}
\email{lsong11@ur.rochester.edu}
\affiliation{%
  \institution{University of Rochester}
  \city{}
  \state{}
  \country{}
}

\author{Ali Vosoughi}
\email{mvosough@ece.rochester.edu}
\affiliation{%
  \institution{University of Rochester}
  \city{}
  \state{}
  \country{}
}

\author{Nguyen Nguyen}
\email{nguyen.nguyen@rochester.edu}
\affiliation{%
  \institution{University of Rochester}
  \city{}
  \state{}
  \country{}
}

\author{Chenliang Xu}
\email{chenliang.xu@rochester.edu}
\affiliation{%
  \institution{University of Rochester}
  \city{}
  \state{}
  \country{}
}
\renewcommand{\shortauthors}{Jing Bi et al.}

\begin{abstract}
The rapid evolution of egocentric video analysis brings new insights into understanding human activities and intentions from a first-person perspective.
Despite this progress, the fragmentation in tasks like action recognition, procedure learning, and moment retrieval, \etc, coupled with inconsistent annotations and isolated model development, hinders a holistic interpretation of video content.
In response, we introduce the EAGLE (Egocentric AGgregated Language-video Engine) model and the EAGLE-400K dataset to provide a unified framework that integrates various egocentric video understanding tasks. 
EAGLE-400K, the \textit{first} large-scale instruction-tuning dataset tailored for egocentric video, features 400K diverse samples to enhance a broad spectrum of tasks from activity recognition to procedure knowledge learning.
Moreover, EAGLE, a strong video multimodal large language model (MLLM), is designed to effectively capture both spatial and temporal information.
In addition, we propose a set of evaluation metrics designed to facilitate a thorough assessment of MLLM for egocentric video understanding.
Our extensive experiments demonstrate EAGLE's superior performance over existing models, highlighting its ability to balance task-specific understanding with holistic video interpretation.
With EAGLE, we aim to pave the way for research opportunities and practical applications in real-world scenarios.
\vspace{-3mm}

\end{abstract}


\maketitle

\section{Introduction}
Understanding human activities and intentions in videos is a key challenge for intelligent systems, requiring advanced reasoning capacities. While there have been advancements in computer vision, the most notable breakthroughs are seen in the evolution of Large Language Models (LLMs)~\cite{openai2023gpt4,chung2022scaling}. 
These models benefit from increased data and model size, resulting in enhanced generalizability, which is often challenging to achieve in computer vision tasks.
By leveraging the pre-trained LLMs~\cite{zheng2023judging,gao2023llama}, MLLMs~\cite{li2023blip,li2022blip,dai2023instructblip,bi2023misar,zhao2023LaViLa,liu2023visual,han2023imagebind,chen2023shikra,hu2022promptcap,hua2024finematch} show superior results to a wide spectrum of multimodal tasks~\cite{goyal2017vqav2,hudson2019gqa,okvqa,mishra2019ocrvqa,schwenk2022okvqa,sidorov2020textcaps,kazemzadeh2014referitgame}. 
Unlike current MLLMs that predominantly focus on images, EAGLE advances to capture spatial and temporal information for more in-depth video analysis.
\begin{figure}[ht!]
    \vspace{-3mm}

    \includegraphics[width=0.425\textwidth]{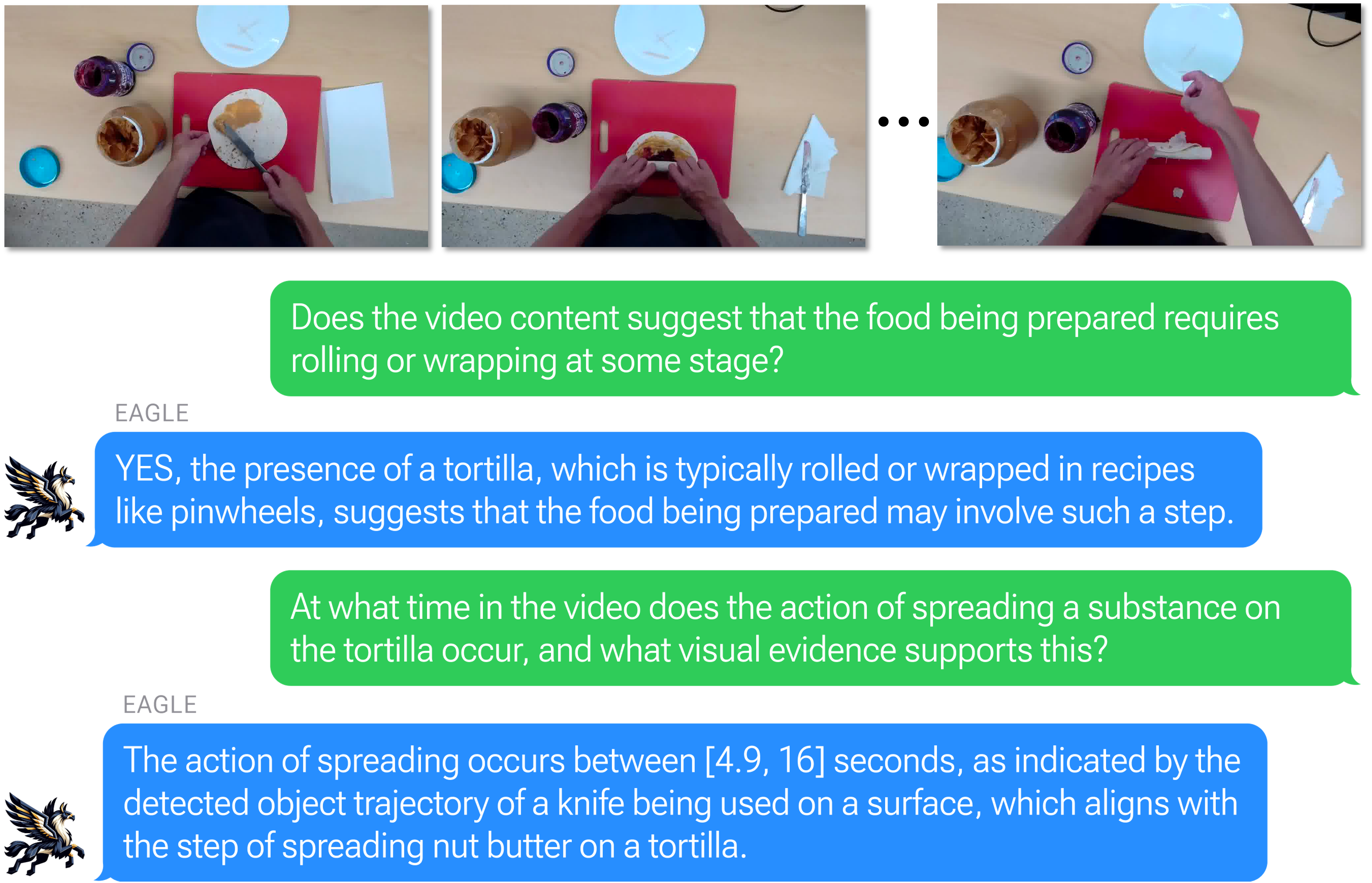}
    \caption{An illustration of EAGLE, a framework designed to unify egocentric video tasks, thereby enhancing both inter-task and intra-task understanding.}
    \Description{This figure illustrates the EAGLE framework, depicting its components and how they interact to unify various egocentric video tasks.}
    \label{fig:teaser}
    \vspace{-2mm}
\end{figure}

To enable MLLM to achieve a more comprehensive and detailed examination of human activities, our work shifts focus from previous efforts centered on third-person perspectives~\cite{kay2017kinetics,soomro2012ucf101,feichtenhofer2019slowfast,caba2015activitynet,kuehne2011hmdb} towards the egocentric view.
This shift not only provides deeper insights into individual interactions with their environment, enhancing tasks such as action recognition and localization, but also enables novel tasks such as Natural Language Queries and Action Anticipation~\cite{grauman2022ego4d}, which demand an in-depth view of the video content, including activity and procedure knowledge understanding.
Taking sandwich preparation as an example, the task involves recognizing actions such as preparing ingredients and spreading condiments, as well as understanding how these actions contribute to the overall process.
Pioneering efforts like EPIC-KITCHENS-100 (EPIC-KITCHENS)~\cite{kazakos2019epic} and Ego4D~\cite{grauman2022ego4d} have paved the way for tasks focused on activities like temporally localizing and anticipating actions. 
Subsequent research~\cite{rhinehart2017first,bi2021procedure,bansal2022my} has extended these concepts by introducing tasks that emphasize procedure knowledge, aiming to understand actions' intentions and contextual relevance.

While a diverse spectrum of tasks offers deeper insights, it also fosters the development of task-specific models. This approach mirrors the traditional methods in NLP, where models specialize in tasks such as sentiment analysis, translation, or question-answering.
For instance, one model may excel at recognizing actions at specific timestamps (e.g., identifying a  \textit{'grab a spoon'} occurring between seconds 5-7), while another model focuses on detecting the precise timing of such actions. 
Although these tasks differ in focus—action recognition versus temporal localization—they both aim to identify the action and its temporal occurrence.
Many works~\cite{jaegle2021perceiver,Huang_2020,Kapidis_2019,Luvizon_2020} have attempted to mitigate these problems by employing a shared backbone~\cite{qinghong2022egocentric,ruder2017overview} or re-scaling labels~\cite{lin2023univtg,xue2023egocentric}.
However, these approaches are limited by their reliance on task-specific models, highlighting the challenge in egocentric video understanding: balancing specialization with a holistic grasp of video content.

Addressing the above challenges, we introduce the EAGLE-400K dataset, the \textit{first} large-scale instruction-tuning dataset designed for egocentric video to advance the understanding of activities and procedure knowledge.
EAGLE-400K comprises 36,000 video clips sourced from Ego4D and EPIC-KITCHENS for activity recognition, as well as PTA for procedural learning. 
By leveraging existing annotations, it facilitates knowledge sharing across datasets, enabling the creation of novel tasks such as Temporal Reasoning and Cross-Referencing Events, which were not present in the original datasets, as detailed in \autoref{tab:video_data_sources} and \autoref{tab:short-example_kitchen}.
Moreover, EAGLE-400K employs instruction-tuning to provide a unified task interface as \texttt{(VIDEO, INSTRUCTION(TASK), RESPONSE)} pairs, thereby serving as a high-quality, large-scale video instruction tuning dataset, as detailed in \autoref{tab:short-example_kitchen}.
Compared with existing visual instruction tuning datasets, such as LLaVA-150K~\cite{liu2023visual} and VideoInstruct100K~\cite{maaz2023video}, our dataset is \textit{3-4$\times$} larger, thereby significantly facilitating research in the field.

Complementing the dataset, we propose EAGLE, a video MLLM, augmenting its capacity for spatial and temporal reasoning through the integration of the Adapter~\cite{houlsby2019parameter}.
We conducted a systematic evaluation to demonstrate the efficacy and adaptability of the proposed dataset and model. By comparing with leading MLLMs, as illustrated in \autoref{fig:radar}, EAGLE outperforms all models on the proposed benchmark.
We summarize our contributions as follows:
\begin{figure}[ht!]
\includegraphics[width=0.48\textwidth]{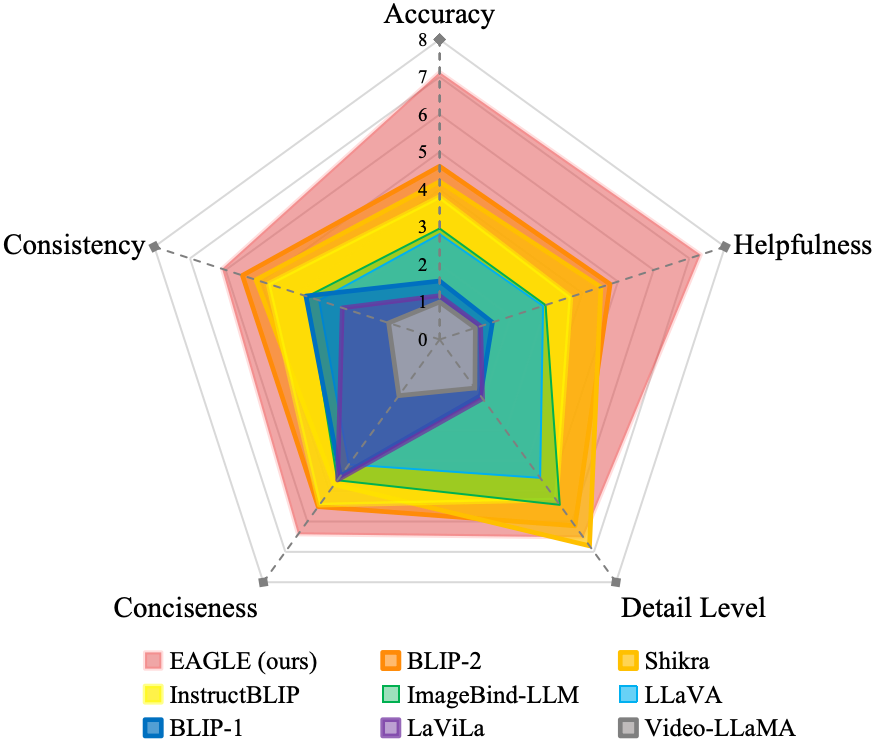}
    \caption{Evaluation results of existing methods, including our EAGLE model and BLIP-2~\cite{li2023blip}, BLIP-1~\cite{li2022blip}, InstructBLIP~\cite{dai2023instructblip}~\etc, using the newly proposed metrics on the EAGLE-400K benchmark. }
    \label{fig:radar}
    \vspace{-3mm}
     \Description{This figure illustrates the EAGLE performace compared with various MLLMs}
\end{figure}
\begin{itemize}[leftmargin=0.1em, noitemsep]
\item \textbf{EAGLE-400K:} The \textit{first} large-scale instruction-tuning dataset designed for egocentric video~\cite{tang2023video}, \textit{4$\times$} the size of the previous largest, is expected to significantly benefit the research community.
\item \textbf{EAGLE Model:} A novel video MLLM designed to incorporate object trajectories, temporal boundaries, and scripted procedure knowledge with the advantage of EAGLE-400K.
\item \textbf{PTA dataset:} To fill the gap in procedural understanding within egocentric view, we have collected and annotated the Perception-driven Task Assistance (PTA) dataset, providing comprehensive and detailed insights into egocentric procedures and interactions.
\item \textbf{Evaluation:} We propose a novel metric and provide a comprehensive assessment of current popular MLLMs, highlighting their limitations in egocentric video understanding.
\end{itemize}

\begin{figure*}[ht!]
  \centering
  \begin{subfigure}[b]{0.7\linewidth}
    \includegraphics[width=\linewidth]{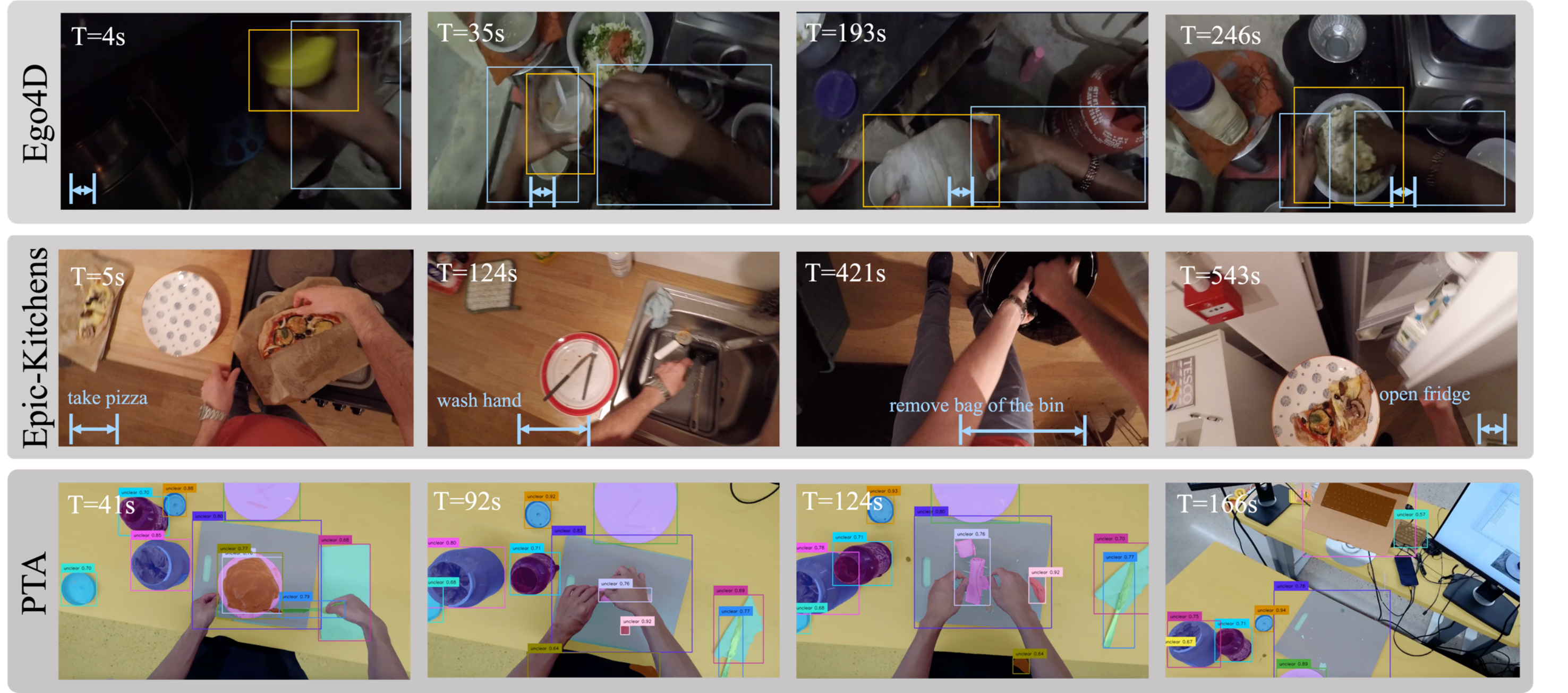}
    \label{fig:source_img}
  \end{subfigure}
  \begin{subfigure}[b]{0.28\linewidth}
    \includegraphics[width=\linewidth]{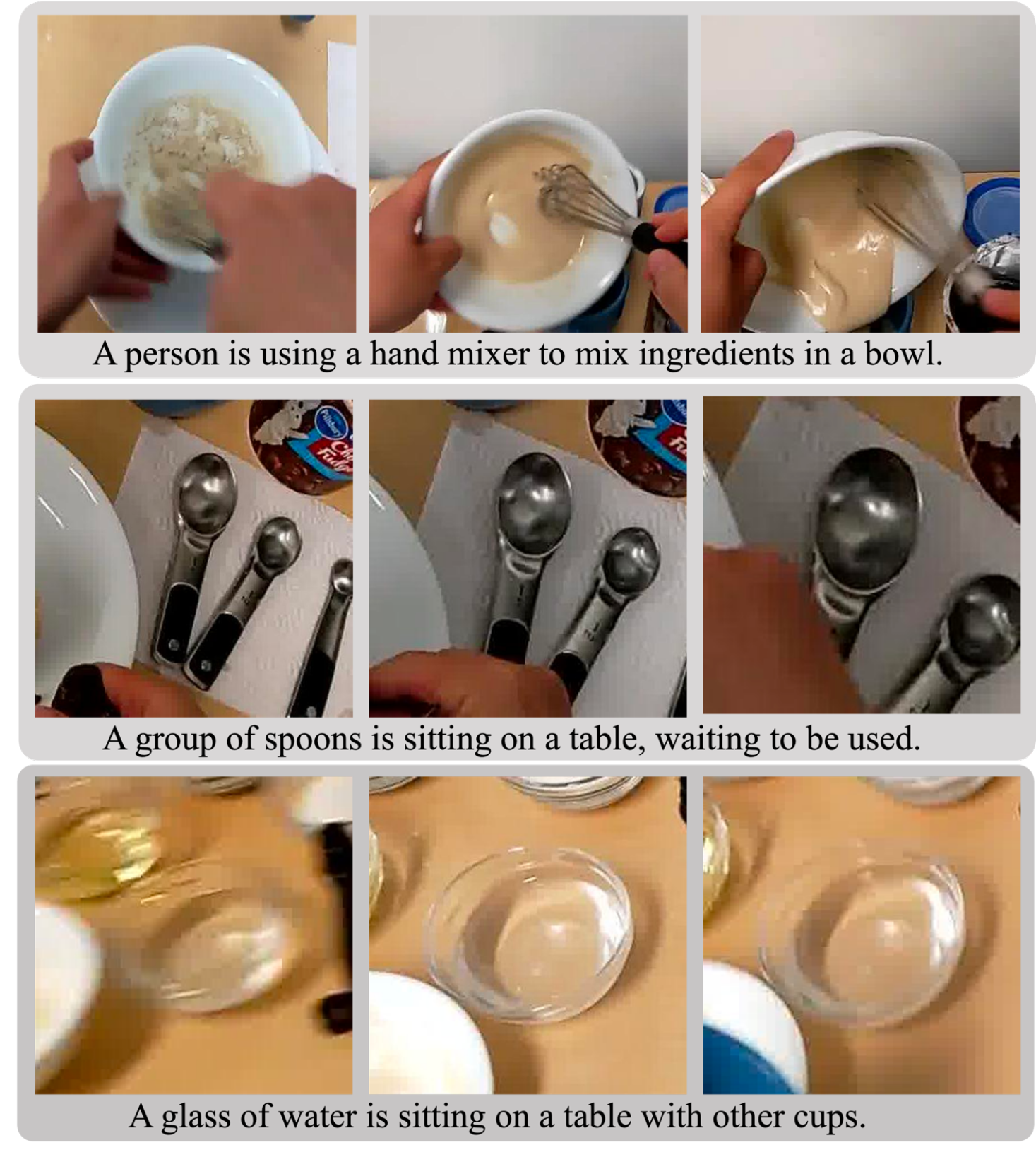}
    \label{fig:ptg_traj}
  \end{subfigure}

  \caption{Left: Representative frames from the Ego4D~\cite{grauman2022ego4d}, EPIC-KITCHENS~\cite{Damen2022RESCALING}, and PTA datasets, showcasing the detailed capture of task-oriented activities. Right: Visualizations of trajectories and object interactions within the EAGLE-400K dataset, emphasizing the tasks' complexity and diversity.}
  \label{fig:combined}
  \vspace{1em}
\end{figure*}
\section{Related Work}
\subsection{Egocentric Video Understanding} Egocentric Video Understanding began with pioneering datasets~\cite{pirsiavash2012detecting,damen2014you,li2015delving} that demonstrated the unique potential of first-person video analysis. The field expanded with EPIC-KITCHENS~\cite{Damen2022RESCALING}, featuring 100 hours of videos, and further with Ego4D~\cite{grauman2022ego4d}, which boasts an impressive 3,000 hours of data. These expansions inspired a wide range of research tasks, including human-object interactions~\cite{nagarajan2019grounded,Xu_2023_ICCV}, activity recognition~\cite{kazakos2019epic,Wen_2023_CVPR,WangHuiyu_2023_ICCV,Radevski_2023_ICCV}, sounding object localization~\cite{HuangCao_2023_CVPR, zhou2015temporal,Akiva_2023_ICCV,Mai_2023_ICCV}, pose estimation and prediction~\cite{Ohkawa_2023_CVPR,Wang_2023_CVPR}, procedure knowledge learning~\cite{Hazra_2023_ICCV,bansal2022my}, and social understanding~\cite{Ryan_2023_CVPR}. 
However, various tasks have resulted in specialized, fragmented model development.
EAGLE-400K aims to consolidate these tasks for a more holistic video understanding.

\subsection{LLMs for Multimodal Understanding} Recent advancements have extended LLMs to multimodal domains, resulting in MLLMs~\cite{li2022blip, zhu2023minigpt, dai2023instructblip, zhang2023video, li2023mimic,hua2024finematch,feng2024more} that excel in various tasks. 
Fine-grained multimodal understanding involves a detailed understanding of visual content, including spatial details~\cite{chen2023shikra,  zhang2023gpt4roi, peng2023kosmos, wang2023caption}, temporal sequences~\cite{lin2023univtg, pan2023scanning, tang2023llmva, Tang_2022_ACCV,tang2024avicuna,hua2024v2xum}, or a combination of both~\cite{bertasius2021space,wang2023all,liu2024emo}.
Models like those in\cite{li2022blip, dai2023instructblip, zhu2023minigpt} use a two-stage Q-former to align vision and language models. \cite{zhang2023video} aligns video and audio modalities with LLMs by training adapters, showing its ability to integrate multiple modalities effectively.
Video-ChatGPT~\cite{maaz2023video} and VideoChat~\cite{li2023videochat}, combining LLMs with video foundation models, are tailored for coarse-grained video-based conversations.
However, few MLLMs are designed to tackle both spatial and temporal video tasks~\cite{tang2023video}, and our work emphasizes interpreting 16 seconds videos, which are \textit{2-4$\times$} longer compared with other video MLLMs.

\subsection{Fine-grained Multimodal Comprehension}
Fine-grained multimodal comprehension involves a detailed understanding of visual content, including spatial~\cite{chen2023shikra,  peng2023kosmos, wang2023caption, xie2023visorgpt}, temporal~\cite{lin2023univtg, pan2023scanning, shang2021multimodal, lin2023univtg, Tang_2022_ACCV}, or both spatial and temporal~\cite{bertasius2021space,wang2023all} information. The multimodal models for fine-grained spatial understanding like \cite{chen2023shikra} and \cite{xuan2023pink} are utilizing LLMs trained on an instruction-tuning dataset which is produced by the language-only GPT-4 and include the coordinates of objects’ bounding boxes. They can handle multiple location-related multimodal tasks like REC, PointQA, dense image captioning, and VQA. In \cite{zhang2023gpt4roi, peng2023kosmos}, special tokens representing the regions are used, while \cite{bai2023qwen} adopts both special tokens and coordinates. \cite{lai2023lisa, wang2023caption} implemented irregular pixel-level region segmentation, generating descriptive captions for any object within an image. The multimodal models for fine-grained temporal video understanding, including~\cite{lin2023univtg, pan2023scanning, tang2023llmva, tang2024avicuna, hua2024v2xum}, are leveraging the capabilities of LLMs. There are seldom models designed to handle both spatial and temporal video understanding tasks.

\begin{table*}[t!]\centering
\setlength\tabcolsep{0pt}
\scalebox{1}{
\makeatletter
\begin{minipage}{1.0\linewidth}\vspace{0mm}    
\centering
\caption{An example task on EAGLE-400K: the preparation of a dish called ``Pinwheel'' from PTA data. It details the dish preparation process and featured objects. The task involves placing various ingredients on a tortilla with a knife, performed by a participant wearing a camera. The top block presents prompts for GPT, including captions and object boxes, while the bottom block shows question types and responses. Notably, the visual image does not prompt GPT.}

\begin{customframed} 
    \centering
    \scriptsize
    \renewcommand{\arraystretch}{1.2}
    \begin{tabular}{p{1\columnwidth} c}
   \VarSty{ {\scriptsize\bf Context type 1: Task Description} } &\\
\scriptsize Pinwheels with steps 1: Place the tortilla on the cutting board., 2: Scoop nut butter and spread it on the tortilla, leaving a margin at the edge., 3: Clean the knife with a paper towel., 4: Using the knife, scoop jelly from the jar and spread it over the nut butter., 5: Clean the knife with a paper towel., 6: Roll tortilla into a tight 1.5-inch thick log without squeezing out the filling, 7: Secure the roll with 5 toothpicks spaced 1 inch apart., 8: Trim tortilla roll ends, leaving a 1/2 inch margin near the last toothpick; discard the ends., 9: Place floss under the roll, halfway between two toothpicks, perpendicular to its length, 10: Cross floss ends over the roll and pull in opposite directions to slice., 11: Continue slicing with floss to create 5 pinwheels., 12: Place the pinwheels on a plate. & \\ \scriptsize The current step, as ground truth, is: $<$0,16$>$ 4: scoop jelly and spread jelly & \\
\VarSty{ {\scriptsize\bf Context type 2: Object Trajectory} } & \\
\scriptsize A jar of ice cream is sitting on a table.: [12, 0.215, 0.57],[4, 0.17, 0.57],[10, 0.2, 0.56],[7, 0.185, 0.545],[6, 0.175, 0.47],[8, 0.175, 0.52],[13, 0.16, 0.53],[14, 0.185, 0.695]
A person is using a knife to spread peanut butter on a plate.: [12, 0.78, 0.61],[9, 0.755, 0.59],[11, 0.75, 0.59],[8, 0.735, 0.555],[6, 0.745, 0.52],[14, 0.73, 0.55],[7, 0.75, 0.57],[10, 0.755, 0.6]
A bowl of peanut butter is being held by a person.: [11, 0.765, 0.68],[0, 0.71, 0.795],[4, 0.75, 0.65],[3, 0.675, 0.645],[13, 0.755, 0.675],[8, 0.735, 0.645],[12, 0.785, 0.705],[14, 0.755, 0.66]
A man is using a knife to cut a piece of cheese on a plate.: [5, 0.79, 0.825],[8, 0.855, 0.81],[6, 0.85, 0.75],[12, 0.875, 0.905],[15, 0.84, 0.94],[14, 0.83, 0.88],[11, 0.905, 0.85],[10, 0.905, 0.895]
A person is holding a bowl of yellow paint.: [6, 0.76, 0.6],[1, 0.87, 0.835],[4, 0.735, 0.63],[10, 0.76, 0.67],[5, 0.725, 0.615],[7, 0.77, 0.645],[11, 0.765, 0.675],[13, 0.74, 0.645]

    \hrulefill & \\
   \VarSty{ {\scriptsize\bf Response type 1: Task Verification} } & \\
\scriptsize\textbf{Question:} Based on the video content, is the person in the video currently spreading something on the tortilla, and if so, what is it? &\\
\scriptsize\textbf{Answer:} YES, the person is spreading jelly on the tortilla. This is indicated by the action of using a knife to scoop out a substance, likely jelly, from a container and then spreading it onto a flat surface, which is consistent with step 4 of the recipe.& \\
   \VarSty{ {\scriptsize\scriptsize\bf Response type 2: Step Verification} }& \\
\scriptsize\textbf{Question:} At what moment in the video can you confirm that jelly is being spread on the tortilla, and what visual clues support this? &\\

\scriptsize\textbf{Answer:} The action of spreading jelly on the tortilla can be confirmed throughout the video, as indicated by the trajectory of the knife moving across the frame, consistent with a spreading motion. The presence of a container from which the jelly is being scooped also supports this. &\\
\scriptsize\textbf{Question:} How does the movement of the knife in the video align with the current step of spreading jelly? &\\
\scriptsize\textbf{Answer:} The movement of the knife in the video aligns with the spreading action required in the current step. The trajectory data shows the knife moving across the frame in a manner consistent with spreading a substance on a flat surface.&\\
   \VarSty{ {\scriptsize\bf Response type 3:  Objects Verification} }& \\
\scriptsize\textbf{Question:} At what time and location in the frame does the jar of jelly appear, and how is it relevant to the task? &\\
\scriptsize\textbf{Answer:} The jar of jelly appears throughout the video, with its location varying slightly but generally centered around the coordinates (0.2, 0.57). It is relevant to the task as it is the source of the jelly being spread on the tortilla, which is a key step in making pinwheels.
    \end{tabular}
\end{customframed}

    \label{tab:example_ptg}
\end{minipage}
\makeatother
}
\end{table*}

\section{EAGLE-400K Dataset and Benchmark}

Egocentric video understanding~\cite{plizzari2023outlook,nunez2022egocentric} involves two primary aspects: \textit{activity recognition}, which identifies actions like picking up objects, and \textit{procedure knowledge learning}, which models the relationships between actions and their contributions to accomplishing tasks.
We aim to \textit{consolidate multiple datasets} with different focuses and provide a comprehensive dataset.
We start with two popular egocentric datasets, EPIC-KITCHENS~\cite{kazakos2019epic} and Ego4D~\cite{grauman2022ego4d}, featuring long-term, untrimmed videos of daily tasks. 
These datasets are annotated with actions labels and object interactions without procedure knowledge, focusing solely on identifying actions.

\begin{table}[b!]
\centering
\caption{
The table compares vision-language instruction-tuning datasets, including EAGLE-400K and MIMIC-IT. MIMIC-IT generates questions from visual descriptions but often produces questions not closely related to the visual content due to noisy narration. VideoInstruct is generated from ActivityNet-200~\cite{7298698} and serves as popular video instruction tuning dataset, featuring short clips paired with QA-style data without spatial-temporal understanding.}
\label{tab:dataset_k}
\scalebox{0.9}{
\begin{tabular}{l|ccccc}
\hlinew{1.15pt}
\textbf{Dataset} & \textbf{Video} & \textbf{\#Clip} & \textbf{\#Ins.} & \textbf{\#Ins./clip} & \textbf{Duration} \\
\hline
MiniGPT-4~\cite{zhu2023minigpt} & $\times$ & - & 5K & - & - \\
Shikra-RD~\cite{chen2023shikra} & $\times$ & - & 5.9K & - & - \\
EgoSchema~\cite{mangalam2023egoschema} & $\checkmark$ & 5k & 3 min \\
LLaVA~\cite{liu2023visual} & $\times$ & - & 345K & - & - \\
MIMIC-IT~\cite{li2023mimic} & $\checkmark$/$\times$ & 400K & 2.4M & 6 & 4-8 frames \\
VideoInstruct~\cite{li2023videochat} & $\checkmark$ & 13k & 100k & 7 & ~5 s \\
\textbf{EAGLE-400K} & $\checkmark$ & 36K & 400K & 11 & 16-76s \\
\hlinew{1.15pt}
\end{tabular}
}
\end{table}

To bridge the gap in procedural understanding we have collected the PTA dataset, including 268 videos recorded in laboratory settings.
This dataset is designed to enhance procedure knowledge learning through three distinct recipes: pinwheel, mug cake, and brew coffee. 
Unlike previous datasets~\cite{bansal2022my,Sener_2022} which prioritized task diversity but lacked depth within individual tasks, our approach focuses on providing extensive variation and a higher number of samples within fewer tasks. 
This approach enables a more comprehensive analysis of procedural steps as shown in \autoref{fig:combined}.

\subsection{Annotation}
We used established training and validation splits for Ego4D and EPIC-KITCHENS. For PTA, we used a 7/3 split, excluding videos from one lab for testing the novel environment, as detailed in \autoref{tab:video_data_sources}. For EPIC-KITCHENS split, we utilized official annotations that include action-object labels with temporal boundaries as shown in \autoref{fig:combined}. 
Additionally, we integrated spatial annotations from the EPIC-KITCHENS-VISOR dataset~\cite{VISOR2022}, an extension of EPIC-KITCHENS, providing object segmentation trajectories covering one-third of the original EPIC-KITCHENS dataset.
In the case of Ego4D, the initial $\sim$3.8 million narrations underwent refinement to generate various subsets, as outlined in \cite{grauman2022ego4d}. Our focus lies on the Episodic Memory and Forecasting Benchmark, which includes tasks such as Natural Language Queries, Moment Queries, and Long-term Action Prediction tasks, all tailored for activity understanding.
In the PTA subset, each video depicts the process of making a recipe, with timestamps marked for key procedure steps.

To enrich the annotation with object information, we first fine-tuned the Grounding DINO~\cite{liu2023grounding} using the EgoObject dataset~\cite{Zhu_2023_ICCV}, omitting its class head. This significantly improved its object proposal accuracy to over \textbf{90\%} on the test set. 
Next, we integrated this specialized Grounding DINO model with the latest DEVA tracker~\cite{Cheng_2023}, achieving reliable object tracking from an egocentric viewpoint.
Lastly, we employed the LLaVA-13B model, known for its robust visual recognizing ability, to interpret the semantic meanings of the proposed object regions.
As shown in ~\autoref{fig:combined}, while this approach may not always achieve the accuracy level of human annotation—occasionally mistaking a tortilla for flatbread—it marks a considerable leap forward, especially given the scarcity of zero-shot vision models capable of high accuracy grounding.

\subsection{Instruction Tuning Data Generation}
\vspace{3mm}

As previously mentioned, diverse tasks and inconsistent annotation standards often limit the comprehensive understanding of videos.
We adapt the instruction tuning~\cite{yin-etal-2023-llm} to unify these annotations under a cohesive framework.
In our dataset, videos are segmented into 16-second clips, \textit{3-5$\times$} longer than common video understanding dataset, ensuring each contains a rich number of actions while maintaining a manageable length, as shown in \autoref{tab:video_data_sources}.
By comparison, our baseline model, LaViLa, which is trained specifically on egocentric videos, typically takes a 1-sec clip. 
Another example is EPIC-KITCHENS Action Anticipation task,  although videos tend to be minutes, only a 5-second segment is used for analysis. 
Adopting 16-second clips allows us to capture comprehensive action details without overwhelming the model.
\begin{table*}[t!]\centering
\setlength\tabcolsep{0pt}
\scalebox{1}{ 
\makeatletter
\begin{minipage}{1.0\linewidth}\vspace{0mm} \centering
\caption{The table outlines activities in a kitchen video, including opening/closing cabinets, grabbing a knife, and washing vegetables, showcasing a person's kitchen work. It serves as an example of data for instruction-following tasks. The top block displays prompts like captions and boxes for GPT, while the bottom block shows response types. Notably, the visual image does not prompt GPT and is included for reference only.}
\begin{customframed} 
    \centering
    \scriptsize
    \renewcommand{\arraystretch}{1.4}
    \begin{tabular}{p{1\columnwidth} c}
   \VarSty{ {\scriptsize\bf Context type 1: Temporal History} } &\\
\scriptsize {Past 30 second}: take container, take tofu, close fridge, open fridge, take carrots, open drawer, close fridge, put down vegetables, open cupboard, take cutting board, put down cutting board
{Current} $<$0,0.76$>$ close cupboard, $<$3.66,5.0$>$ open drawer,$<$5.5,8.0$>$ take knife, $<$5.55,6.36$>$ take knife, $<$6.73,7.74$>$ put down knife, $<$8.1,8.6$>$ close drawer, $<$9.45,10.7$>$ put down knife, $<$11.05,14.3$>$ open tap,$<$13.86,16$>$ wash courgette & \\
\scriptsize {Future 30 second}: still washing courgette, wash carrot, still washing carrot, close tap, put down vegetables, open cupboard, take grater, take pan, put down pan, close cupboard, take courgette, cut courgette& \\

\VarSty{ {\scriptsize\bf Context type 2: Object Trajectory} } & \\
\scriptsize 'right hand':
[[5.0, 0.295, 0.401], [6.0, 0.317, 0.419], [7.0, 0.294, 0.365], [8.0, 0.324, 0.406], [10.0, 0.303, 0.377], [12.0, 0.344, 0.366], [13.0, 0.336, 0.284]]
'left hand':
[[5.0, 0.186, 0.407], [6.0, 0.207, 0.427], [7.0, 0.212, 0.402], [8.0, 0.214, 0.402], [11.0, 0.293, 0.368], [12.0, 0.113, 0.324], [13.0, 0.216, 0.358]]
'drawer':
[[5.0, 0.258, 0.311], [6.0, 0.193, 0.319], [7.0, 0.227, 0.284], [8.0, 0.219, 0.31]]
'chopping board':
[[9.0, 0.266, 0.108], [10.0, 0.203, 0.348], [11.0, 0.153, 0.373], [12.0, 0.045, 0.395], [13.0, 0.035, 0.404]]
'carrot':
[[10.0, 0.218, 0.177], [11.0, 0.241, 0.233], [12.0, 0.096, 0.252], [13.0, 0.075, 0.264]]

    \hrulefill & \\
   \VarSty{ {\scriptsize\bf Response type 1: Description} } & \\
\scriptsize \textbf{Question:} Can you provide a concise summary of the main events occurring in the video? &\\
\scriptsize \textbf{Answer:} Certainly.  The video shows a sequence of kitchen activities, including opening and closing a cupboard and drawer, taking and putting down a knife, and washing a courgette. & \\
   \VarSty{ {\scriptsize\bf Response type 2: Detailed Description} }& \\
\scriptsize \textbf{Question:} Can you provide a detailed account of the sequence of events that take place between 3.66 and 8.6 seconds in the video, specifically focusing on the actions involving the kitchen drawer and knife? Please include the timing of these actions and describe how the person's hands interact with these objects, as indicated by the object trajectory data.&\\
\scriptsize \textbf{Answer:} Between 3.66 and 8.6 seconds in the video, the individual engages in a sequence of actions involving a kitchen drawer and a knife. Initially, between [3.66, 5.0] seconds, the right-hand reaches for the drawer, with its trajectory indicating a movement slightly right and above the center, coinciding with opening the drawer. Then, the person takes a knife between [5.5, 6.36] seconds, as evidenced by the right hand moving downward and leftward towards the drawer. This action involves a brief re-adjustment, adjusts their grip, and grasps it again. There's a moment between [6.73, 7.74] seconds where the knife is put down, suggested by the right hand moving away from the drawer. Finally, the drawer is closed between [8.1, 8.6] seconds, with the right hand's trajectory returning to its original position, indicating the closing motion. Throughout this sequence, the left hand appears to provide support or stabilization, with minor adjustments in its position.
&\\
   \VarSty{ {\scriptsize\bf Response type 4: Event Localization} }& \\
\scriptsize \textbf{Question:} At what timestamps does the individual open and close the drawer?&\\
\scriptsize \textbf{Answer:} The individual opens the drawer between [3.66, 5.0] seconds and closes it between [8.1, 8.6] seconds in the video.
&\\
   \VarSty{ {\scriptsize\bf Response type 5: Temporal Reasoning} }& \\
\scriptsize \textbf{Question:} How did the previous actions of taking vegetables and opening the fridge lead to the current situation in the video?&\\
\scriptsize \textbf{Answer:} Previous actions like taking vegetables and opening the fridge led to the current situation by indicating food preparation activities, culminating in washing the courgette.
&\\
   \VarSty{ {\scriptsize\bf Response type 6: Event Localization} }& \\
\scriptsize \textbf{Question:} Given the ongoing event of washing vegetables, what might be a likely future action sequence in the kitchen, and why?&\\
\scriptsize \textbf{Answer:} Based on the ongoing event of washing vegetables, future actions may include chopping the courgette, possibly using the grater and pan (as indicated in the 'future' events list), suggesting a continuation of food preparation.
&\\
   \VarSty{ {\scriptsize\bf Response type 7: Cross-Referencing Events} }& \\
\scriptsize \textbf{Question:} Can you explain the connection between opening the drawer (3.66-5.0 seconds) and the subsequent use of the chopping board (after 9.0 seconds)?&\\
\scriptsize \textbf{Answer:} The opening of the drawer (3.66-5.0 seconds) and the use of the chopping board (after 9.0 seconds) are connected as both actions are part of setting up for the food preparation process; utensils are gathered first (from the drawer), followed by setting up the chopping board for cutting vegetables.
    \end{tabular}
\end{customframed}
\vspace{2mm}

    \label{tab:short-example_kitchen}
\end{minipage}
\makeatother
}
\end{table*}

To determine the optimal frame rate, we draw inspiration from recent studies~\cite{rasheed2023fine, xu2021videoclip} that have shown promising results in frame-based video understanding by analyzing videos frame-by-frame and using feature pooling. 
Building on this, we sample one frame per second, maintaining a consistent interval regardless of the video's frame rate. 
To enhance contextual understanding, we incorporate temporal context with 30 seconds before and after each clip. 
We chose a 30-second duration to balance action details and cohesive narration. This is based on our observation that longer durations reduce the relevance of actions.

In this way, the context helps tasks like action anticipation and detection and encourages the development of new tasks by extrapolating relationships between labels. 
For instance, our framework enables advanced tasks such as Temporal Reasoning and Cross-Referencing Events, as shown in \autoref{tab:short-example_kitchen}, enhancing the dataset's utility without additional annotation effort.

We use two types of symbolic representations to prompt GPT4: (i) Captions, which typically describe the visual scene from various perspectives. (ii) Object trajectories in the scene, and each box encodes the object concept and its spatial location as shown in \autoref{fig:combined}.
We collect 400K unique video instruction-following samples in total, including 350K for activity recognition (as shown in \autoref{tab:short-example_kitchen}) and 50K for procedure knowledge learning.
We have undertaken multiple iterations to refine our method for creating accurate instruction data from task descriptions and object trajectories. 
We normalized object bounding boxes to a scale of 0-1 and used only the center points of objects, improving the spatial relationships in GPT-4's responses. 
Additionally, we added a post-processing step that uses interpolation to align GPT-4 output coordinates with actual object trajectories, ensuring high data accuracy. 
However, including complete trajectories in responses sometimes led to errors. 
To counter this, we selectively replaced faulty segments with ground truth data, enhancing the dataset's usability.

As shown in \autoref{tab:dataset_k}, our approach provides longer question-to-clip correspondence than MIMIC-IT~\cite{li2023mimic}, focusing on video content comprehension. In contrast, MIMIC-IT~\cite{grauman2022ego4d} often generates questions unrelated to the visual content. Compared to EgoSchema~\cite{mangalam2023egoschema}, our method emphasizes fine-grained understanding, while EgoSchema targets coarse-grained analysis with few multiple-choice questions for 3-minute videos.

\section{EAGLE Model}
\begin{figure*}[ht!]
    \centering
    \vspace{2mm}
    \includegraphics[width=0.985\linewidth]{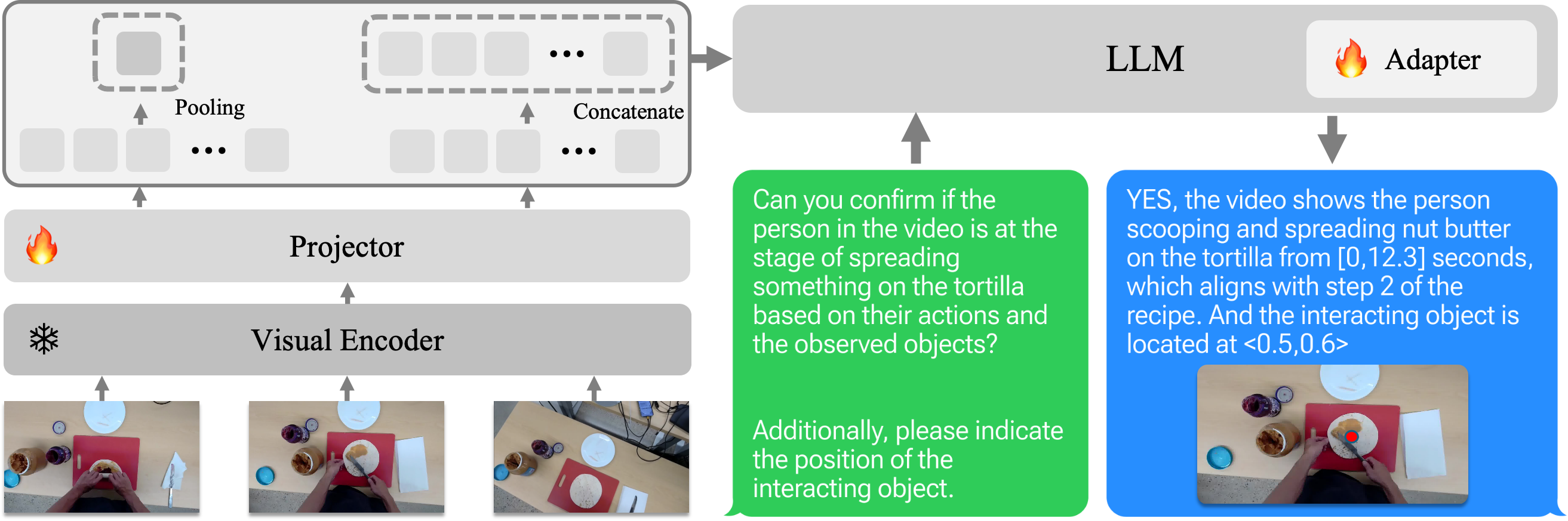}
    \caption{The architecture of the EAGLE model, which includes a finely-tuned projection layer and adapter, enhancing the language model’s capability to handle complex instructions containing temporal boundaries and object location tokens. This design enables the model to accurately determine the precise temporal boundaries of events and identify the specific locations of objects within a given context.}
    \vspace{2mm}
    \label{fig:model}
\end{figure*}

Existing image-based MLLMs such as Shikra~\cite{chen2023shikra} primarily focus on spatial information, while models like VTimeLLM~\cite{huang2023vtimellm} specifically target temporal dimensions.
Given the unique aspects of our dataset, which encompasses both spatial and temporal attributes, our goal is to simplify the tuning process and construct a straightforward yet strong model by leveraging the existing MLLM model.

Our model, in line with common MLLMs, integrates a vision encoder, an alignment layer, and a large language model (LLM), specifically employing the pre-trained ViT-L/14 from CLIP~\cite{radford2021learning} as the frame encoder $\mathbb{E}$ and Vicuna-13B as LLM, as shown in \autoref{fig:model}.
Given a video sample $V_i \in \mathbb{R}^{T \times H \times W \times C}$ with $T$ frames, the frame encoder $\mathbb{E}$ processes each frame independently, generating video embedding as $x_i \in \mathbb{R}^{T \times D}$. 
After obtaining frame embeddings, selecting an optimal method for aggregating these features is critical.

Video-LLaMA~\cite{zhang2023videollama} employs temporal position embedding and a q-former, which typically demands a large amount of paired video-text data (rare in video datasets).
Compared with image-language datasets such as CC3M~\cite{sharma-etal-2018-conceptual} utilized by LLaVA~\cite{liu2023improved},  video-language datasets like WebVid~\cite{Bain_2021} contain shorter and less detailed language descriptions.
Consequently, when models are pretrained on these video datasets, their expressiveness is often limited, which can result in a less effective image-language alignment layer.
Instead, we choose to leverage existing alignment layers from LLaVA to obtain language tokens from visual features.
We have two strategies, (i) adapt recent advancements \cite{rasheed2023fine} and employ an average-pooling strategy to aggregate a video-level representation $v_i \in \mathbb{R}^{D}$, where D is 5,120 for Vicuna-13B. We denote this model as EAGLE-pool.
(ii) Instead of using pooling, we employ alignment layers to extract language tokens directly from each visual frame and concatenate these tokens into a long sequence. This method does not require the explicit embedding of position tokens. Instead, it implicitly incorporates temporal learning, thus leveraging the strengths of the LLaVA alignment layer, which ensures more reliable alignment compared to Q-former aggregation methods.

To enhance the LLM's ability to capture both temporal and spatial information, we integrated Adapters~\cite{houlsby2019parameter} into various self-attention layers of Vicuna-13B, allowing the model to effectively incorporate coordinates from both time boundaries and object trajectories.
During training, the visual embedding can be inserted anywhere in the input sequence.
Regarding the frame encoder $\mathbb{E}$, we decided to keep the visual encoder frozen throughout all training phases, as fine-tuning the visual encoder even with a small-scale dataset can affect its image representation capabilities and yield performance drop, as discussed in \cite{wang2023makes}. 

Following~\cite{liu2023visual,zhang2023videollama}, the model training is done in two phases. In the first phase, we only focus on fine-tuning the projection layer with a subset of \texttt{(VIDEO, INSTRUCTION, RESPONSE)} pairs that do not include time boundary and object trajectory. 
This stage is primarily focused on aligning visual tokens with LLM text tokens, minimizing distractions from numerical inputs.
In the second phase, both the newly integrated Adapters and the projection layer are trained using the entire dataset on 8 NVIDIA A100 GPUs. 
This stage aims to enhance the model’s performance on instruction tasks involving temporal and spatial reasoning.
Our model establishes a robust baseline and sets the stage for future research into more accurate temporal-spatial grounding abilities and context modeling.

\section{Experiments}
\begin{table*}[h!]
  \centering
  \caption{Video sources and the corresponding number of videos and average actions for training and validation sets.}
\scalebox{1.1}{
\begin{tabular}{l|cc|cc|c}
\hlinew{1.15pt}
\multirow{2}{*}{Video sources} & \multicolumn{2}{c|}{Training set}                 & \multicolumn{2}{c|}{Validation set}               & \multirow{2}{*}{Total} \\ \cline{2-5}
                               & \multicolumn{1}{c|}{\# videos} & \# actions (avg) & \multicolumn{1}{c|}{\# videos} & \# actions (avg) &                        \\ \hline
EPIC-KITCHENS~\cite{Damen2022RESCALING}                  & \multicolumn{1}{c|}{16,570 (57\%)}     & 4.78             & \multicolumn{1}{c|}{2,901 (38\%)}      & 3.98             & 19,471 (53\%)                 \\
Ego4D~\cite{grauman2022ego4d}                          & \multicolumn{1}{c|}{9,050 (31\%)}      & 2.30             & \multicolumn{1}{c|}{3,669 (47\%)}      & 2.80             & 12,719 (35\%)                 \\
PTA                           & \multicolumn{1}{c|}{3,355 (12\%)}      & 1.55             & \multicolumn{1}{c|}{1,167 (15\%)}      & 1.53             & 4,522 (12\%)                  \\ \hline
Total                          & \multicolumn{2}{c|}{28,975}                        & \multicolumn{2}{c|}{7,737}                         & 36,712                  \\ \hlinew{1.15pt}
\end{tabular}
}
  \label{tab:video_data_sources}
\end{table*}

\subsection{Evaluation Metrics} 
Following the evaluation methods~\cite{liu2023gpteval,zheng2023judging} for recent LLMs, we use GPT-4 to assess the quality of responses generated by models. Due to the time-consuming nature of evaluating all 7,700 samples across nine models with GPT-4, we adopt a square root sampling strategy, selecting approximately (\(\sqrt{7700} \approx 88\)) 100 samples as a representative subset.
To maintain consistency and ensure the reproducibility of findings from the initial 100 samples, we further analyzed 200 additional responses. This was done to evaluate the performance of the top four models, which we have designated as EAGLE-pool$_2$, Shikra$_2$, BLIP-2$_2$, and EAGLE$_2$. The results are presented in \autoref{table:result}.
The results from this extended dataset are presented in the subsequent table and are consistent with the findings from our initial sample of 100 responses.

Given the nature of the egocentric dataset, which offers only action labels, recipe steps, and corresponding timestamps, we need to develop ground truth sentences for evaluation purposes. Our empirical findings indicate that compared to using polished sentences of ground truth labels, template-based construction reduces the occurrence of hallucination errors. The evaluation prompt was refined iteratively through trial and error, aiming to improve the accuracy in identifying event boundaries and objects and to enhance clarity. The evaluation prompt will be included in the supplementary.

These selected responses will be scored by GPT-4 based on five key metrics, each rated on a scale from 1 to 10, with higher scores indicating superior performance. The evaluation metrics are as follows:
\begin{enumerate}[leftmargin=2em,noitemsep]
\item \textit{Accuracy}: This metric involves assessing if the response reflects the video's content, focusing on activity recognition for EPIC-KITCHENS and Ego4D samples, and the match between predicted and ground truth procedure steps for PTA samples. 

\item \textit{Helpfulness}: evaluating how much the response aids in comprehending the video's content and its broader context. It involves assessing whether the model's output provides actionable insights or clarifies complex elements within the video.

\item \textit{Level of Detail}: This involves assessing the comprehensiveness and specificity with which the video is described. A high score in this area indicates that the model captures essential objects and events of the video.

\item \textit{Conciseness}: This metric measures the succinctness and clarity of the response, focusing on delivering essential information without superfluous content. Effective conciseness involves distilling complex information into a clear and brief explanation, which is critical for providing essential information of the video.

\item \textit{Consistency}: This assesses the uniformity and reliability of the narrative or description provided by the model across multiple instances or parts of the video. 
\end{enumerate}
Please note that the descriptions provided above are instruction prompts for GPT-4. 
Metrics such as accuracy and detail assess the alignment between the outputs and the established ground truth, including the accurate representation of objects. Subjective metrics like helpfulness and conciseness focus on the quality of the language, ensuring that the responses aid users in grasping the broader context and intent of scenarios.

\subsection{Baseline Models} For our baseline models, we use both image-based and video-based approaches. Image-based models include:

\begin{enumerate}[leftmargin=2em,noitemsep]
\item \textit{BLIP-1}~\cite{li2022blip}, image-language pre-training model that integrates textual and visual information to enhance multimodal understanding.  This model excels in multimodal understanding and is effective in zero-shot video language tasks. 

\item \textit{BLIP-2}~\cite{li2023blip} trained a lightweight Q-Former for multimodal representation alignment and vision-to-language generation, capable of following instructions without multimodal instruction tuning.

\item \textit{InstructBLIP}~\cite{dai2023instructblip}, built upon BLIP-2, this model reformats 26 public datasets for instruction tuning and updates only the Q-Former during training. It formulates various tasks as instructions, similar to our method.

\item \textit{LaViLa}~\cite{zhao2023LaViLa} is a video narration method that pairs a video encoder with a GPT-2~\cite{Radford2019LanguageMA} as language decoder and a T-5~\cite{raffel2020exploring} to reduce overfitting and enhance natural language data.

\item \textit{LLaVA}~\cite{liu2023visual} introduces visual instruction tuning, using GPT-generated data and instructions for conversation, detailed description, and complex reasoning.

\item \textit{ImageBind-LLM}~\cite{han2023imagebind} is an open-source MLLM, with its algorithm details pending publication.

\item \textit{Shikra}~\cite{chen2023shikra} encodes regions in natural language as numerical coordinates to specify regions in user queries.

\item \textit{Video-LLaMA}~\cite{zhang2023video} trains adapters for aligning video and audio modalities with LLMs, sampling only eight frames from arbitrarily long videos.
\end{enumerate}

Among baseline models, LaViLa is specifically trained on egocentric videos (Ego4D, EPIC-KITCHENS) to generate narrations. 
Despite this targeted training, our research reveals that in zero-shot learning scenarios, MLLM outperformed LaViLa for handling egocentric data.
Details of the responses from different models will be included in the supplementary material.

Additionally, to ensure a fair comparison, we chose not to fine-tune the vision encoder in our EAGLE model for egocentric vision adaptation.  Instead, we focused on refining the model to improve its spatial-temporal video analysis capabilities. Our findings indicate that our dataset significantly contributes to enhancing the performance of current MLLMs in understanding and interpreting video content. 
\begin{table*}[!h]
	\centering
	\caption{We evaluated existing models and our EAGLE model. The scores reflect the models' performance in key aspects, with EAGLE achieving the highest scores in Accuracy and Helpfulness, and competitive scores in other areas. Higher scores indicate better performance.}
	\scalebox{1}{
		\begin{tabular}{l|ccccc|c}
			\hlinew{1.15pt}
			\textbf{Model}                          & \multicolumn{1}{l}{\textbf{Accuracy}} & \multicolumn{1}{l}{\textbf{Helpfulness}} & \multicolumn{1}{l}{\textbf{Detail}} & \multicolumn{1}{l}{\textbf{Conciseness}} & \multicolumn{1}{l|}{\textbf{Consistency}} & \multicolumn{1}{l}{\textbf{Average}} \\ \hline
			Video-LLaMA~\cite{zhang2023video}       & 1.00                                  & 1.00                                     & 1.60                                & 1.85                                     & 1.43                                      & 1.38                                 \\
			LaViLa~\cite{zhao2023LaViLa}            & 1.17                                  & 1.15                                     & 1.95                                & 4.63                                     & 2.73                                      & 2.33                                 \\
			BLIP-1~\cite{li2022blip}                & 1.56                                  & 1.48                                     & 1.85                                & 4.50                                     & 3.75                                      & 2.63                                 \\
			LLaVA~\cite{liu2023visual}              & 2.81                                  & 2.9                                      & 4.56                                & 4.12                                     & 3.38                                      & 3.55                                 \\
			ImageBind-LLM~\cite{han2023imagebind}   & 2.96                                  & 2.97                                     & 5.45                                & 4.64                                     & 3.71                                      & 3.95                                 \\
			InstructBLIP~\cite{dai2023instructblip} & 3.81                                  & 3.68                                     & 5.29                                & 5.46                                     & 4.81                                      & 4.61                                 \\
			Shikra~\cite{chen2023shikra}            & 4.21                                  & 4.52                                     & 6.80                                & 4.78                                     & 5.15                                      & 5.09                                 \\
			
			Shikra$_2$                              & 4.31                                  & 4.55                                     & \ul{6.85}                        & 4.20                                     & 5.20                                      & 5.02                                 \\ 
			BLIP-2~\cite{li2023blip}                & 4.62                                  & 4.78                                     & 6.14                                & 5.51                                     & 5.53                                      & 5.32                                 \\ 
			
			BLIP-2$_2$                              & 4.43                                  & 4.80                                     & 6.20                                & 5.45                                     & 5.38                                      & 5.25                                 \\   \hline
			
			\textbf{EAGLE-pool}                     & 7.13                                  & 7.32                                     & 6.52                                & \ul{6.45}                             & 6.10                                      & 6.70                                 \\ 
			\textbf{EAGLE-pool}$_2$                 & \ul{7.21}                          & \ul{7.40}                             & 6.72                                & 6.42                                     & \ul{6.30}                              & \ul{6.81}                         \\ 
			
			\textbf{EAGLE}                          & \textbf{7.32}                         & \textbf{7.51}                            & \textbf{6.90}                       & \textbf{6.75}                            & 6.65                                      & \textbf{7.03}                        \\ 
			
			\textbf{EAGLE$_2$}                      & 7.28                                  & 7.48                                     & 6.83                                & 6.67                                     & \textbf{6.77}                             & 7.01                                 \\ \hlinew{1.15pt}
		\end{tabular}
	}
	\label{table:result}
\end{table*}

\subsection{Results and Analysis}To validate the performance of EAGLE, we compare it with recent MLLMs~\cite{liu2023visual, chen2023shikra, zhang2023video}, on the EAGLE-400K dataset. As \autoref{table:result} shows, Shikra and BLIP-2 demonstrate remarkable proficiency, scoring highest in most categories, indicating their reliability, helpfulness, and detailed response capability. 
Although Video-LLaMA is targeted at video analysis, it exhibits the lowest performance when compared to image-based multimodal large language models (MLLMs), with outputs often arbitrary and failing to capture essential visual information from videos. LLaVA and InstructBLIP demonstrate balanced and above-average performances across all metrics, showcasing their versatility in handling diverse tasks.

Interestingly, while LaViLa is specifically trained on egocentric data, its performance is hindered by its relatively weaker language backbone (GPT-2), resulting in it being outperformed by more advanced MLLMs in a zero-shot setting. This highlights the significant impact that a robust language model can have on performance.

Moreover, ImageBind-LLM excels in providing detailed and consistent responses. This suggests that superior language modeling capabilities, coupled with a more generalized visual encoder, can enhance overall performance significantly.

Comparing the two variants of EAGLE, which utilize different methods for processing video content:
Using concatenation of frame features preserves the temporal order of each frame, allowing the model to capture more detailed temporal dynamics and intricate interactions within the video content. EAGLE-pool, on the other hand, employs temporal pooling to aggregate features over time. This approach helps reduce the impact of less relevant information and noise but may also gloss over finer temporal details that are crucial for understanding complex dynamics. Despite these trade-offs, EAGLE-pool still benefits from the extensive EAGLE-400K dataset and performs better than spatial grounding models like Shikra, which focuses more on spatial rather than temporal data.

\noindent \textbf{Ablation Study.}
Studies were conducted on the EAGLE-400k dataset using varied training data splits to investigate the impact of spatial and temporal information on egocentric video understanding. The ablation included: removing time boundaries (\textit{w/o time}), excluding object trajectories (\textit{w/o obj}), and eliminating both (\textit{only desc}). As shown in Table~\ref{tab:ab}, performance tends to decrease when either time or object information is excluded, with the least effective results observed when relying solely on descriptions. Notably, PTA exhibits the most significant decline in performance when detailed information is removed, indicating that procedural learning relies more heavily on temporal and object details.

The results from Table~\ref{tab:ab} highlight specific trends across different datasets. For EPIC-KITCHENS, excluding temporal information resulted in a performance drop from 6.8 to 5.9, showing a considerable dependence on time data. Similarly, Ego4D saw a decrease from 6.4 to 6.1 and 6.2 without time and object information, respectively. The PTA dataset showed a marked drop from 6.5 to 5.8 when object information was excluded, underscoring its reliance on object trajectories. These findings underscore the critical role of temporal and object-based features in enhancing egocentric video understanding, with the integrated use of all information sources yielding the highest performance across all datasets.
\begin{table}[!ht]
	\centering
	\caption{Ablation study with the different split of the dataset}
	\label{tab:ab}
	\scalebox{1}{
		\begin{tabular}{l|ccc}
			\hline
			\textbf{Dataset} & \textbf{EPIC-KITCHENS} & \textbf{Ego4D} & \textbf{PTA} \\
			\hline
			w/o time & 5.9 & 6.1 & 5.9 \\
			w/o object & 6.2 & 6.2 & 5.8 \\
			only desc & 5.5 & 5.8 & 5.5 \\
			all & 6.8 & 6.4 & 6.5 \\
			\hline
		\end{tabular}
	}
\end{table}
\section{Conclusion}
In this work, we present the EAGLE-400K dataset and the EAGLE model for holistic egocentric video understanding. The EAGLE-400K dataset consists of 40K question-answer pairs from 36K diverse video clips and EAGLE offers a unified framework for diverse visual computational tasks. We also provide an evaluation method for egocentric vision tasks and demonstrate EAGLE's superior performance. The introduction of a new evaluation metric enhances the understanding of video-based MLLMs. 
We hope our work can pave the way for augmented reality assistants that aid in complex physical tasks with multimodal perception.
\section{Limitation}
Our PTA dataset was assembled with a significantly smaller number of contributors compared to larger datasets like EPIC-Kitchens and Ego4D. This limited participant pool may result in the dataset predominantly reflecting individual-specific characteristics, such as a participant’s height or unique culinary techniques, which could skew the representativeness of the data. Additionally, the dataset primarily comprises cooking videos. This focus was chosen because these activities align well with structured instructions and are more readily accessible for recording. However, this emphasis on cooking-related content may introduce a domain bias, as it limits the diversity of egocentric experiences captured, notably underrepresenting categories such as social interactions or spontaneous activities.
\clearpage
\newpage

\section{Acknowledgments}
We express our profound gratitude to Yayuan Li, Filippos Bellos, Professor Jason J. Corso, Yuwei Bao, Shane Storks, and Professor Joy Chai from the University of Michigan for their meticulous efforts in recording and annotating data. 
Special thanks are extended to Juan Carlos and Professor Enrique Dunn from Stevens Institute of Technology for their exceptional work in developing the HL2ss~\cite{dibene2022hololens} technology, enabling us to utilize the HoloLens in our research effectively.
We are deeply appreciative of Professor Jeffrey Mark Siskind from Purdue University for his support in the experiment setup and for providing technical insights that significantly enhanced our study.
We would also like to thank Kalsey Colotl and Miranda Rublaitus for their assistance in recording and annotating data as part of the REU program, which greatly facilitated our research.

This work was supported by the Defense Advanced Research Projects Agency (DARPA) under the PTG Program, Contract No. HR00112220003 and the National Institutes of Health (NIH) under R01EY034562. This paper does not necessarily reflect the position of the Government, and no official endorsement should be inferred.
\vspace{-6mm}
\bibliographystyle{acm}
\bibliography{main}

\end{document}